\documentclass[10pt,twocolumn,letterpaper]{article}

\usepackage{iccv}
\usepackage{times}
\usepackage{epsfig}
\usepackage{graphicx}
\usepackage{amsmath}
\usepackage{amssymb}
\usepackage{booktabs}

\usepackage[breaklinks=true,bookmarks=false]{hyperref}
\usepackage{xcolor}

\iccvfinalcopy 

\def\iccvPaperID{****} 

\ificcvfinal\fi

\definecolor{mediumpersianblue}{rgb}{0.0, 0.4, 0.65}
\definecolor{linkcolor}{RGB}{150, 50, 50}
\hypersetup{colorlinks,linkcolor={linkcolor},citecolor={mediumpersianblue},urlcolor={mediumpersianblue}} 
\begin{document}

\title{Bag of Tricks for Long-Tailed Multi-Label Classification on Chest X-Rays}


\author{Feng Hong\textsuperscript{1,$\ast$} \quad Tianjie Dai\textsuperscript{1,$\ast$} \quad Jiangchao Yao\textsuperscript{1,2} \quad Ya Zhang\textsuperscript{1,2}
\quad Yanfeng Wang\textsuperscript{1,2} \\
\textsuperscript{1}Cooperative Medianet Innovation Center, Shanghai Jiao Tong University \quad
\textsuperscript{2}Shanghai AI Laboratory \quad\\
\tt\small{\{feng.hong, elfenreigen, Sunarker, ya\_zhang, wangyanfeng\}@sjtu.edu.cn}\\
}

\maketitle
\ificcvfinal\thispagestyle{empty}\fi

\begin{abstract}

Clinical classification of chest radiography is particularly challenging for standard machine learning algorithms due to its inherent long-tailed and multi-label nature. However, few attempts take into account the coupled challenges posed by both the class imbalance and label co-occurrence, which hinders their value to boost the diagnosis on chest X-rays (CXRs) in the real-world scenarios. Besides, with the prevalence of pretraining techniques, how to incorporate these new paradigms into the current framework lacks of the systematical study. This technical report presents a brief description of our solution in the ICCV CVAMD 2023 CXR-LT Competition. We empirically explored the effectiveness for CXR diagnosis with the integration of several advanced designs about data augmentation, feature extractor, classifier design, loss function reweighting, exogenous data replenishment, etc. In addition, we improve the performance through simple test-time data augmentation and ensemble. Our framework finally achieves 0.349 mAP on the competition test set, ranking in the top five. 


\end{abstract}


\begingroup\renewcommand\thefootnote{\textsuperscript{$\ast$}}
\footnotetext{Equal Contributions}
\endgroup

\section{Introduction}
Deep learning has been applied to a broad range of medical imaging tasks with remarkable success~\cite{erickson2017machine,titano2018automated}, driving medical imaging understanding into the new era. However, when it comes to the real-world scenarios, the performance of current deep learning methods on X-rays, computed tomography (CT) scans, magnetic resonance imaging (MRI), positron emission tomography (PET) scans, etc., is still far from practical expectation. Especially, when there is a significant imbalance in the training data, machine learning algorithms tend to favor common diseases at the expense of performance in rare diseases, leading to the fairness and safety concerns about the dramatic performance gaps across diseases~\cite{holste2022long, zhang2023deep}. In addition, label co-occurrence often exists in medical diagnosis, as patients might present with multiple diseases simultaneously, which is different from the conventional multi-class setting of machine learning~\cite{chen2020label}. Existing efforts have independently studied the long-tailed learning problem~\cite{cui2019class,hong2023long, zhou2022contrastive} and the multi-label learning problem~\cite{allaouzi2019novel, chen2020label}, but few of them have paid attention to the combined challenges brought by class imbalance and label co-occurrence~\cite{holste2022long}. In this technical report, we focus on the long-tailed multi-label classification problem on chest X-rays and present the solution of our framework design in ICCV CVAMD 2023 CXR-LT Competition.

\section{Method}
\subsection{Problem Formulation and General Framework}
\begin{figure*}[t]
    \centering
    \includegraphics[width=0.7\textwidth]{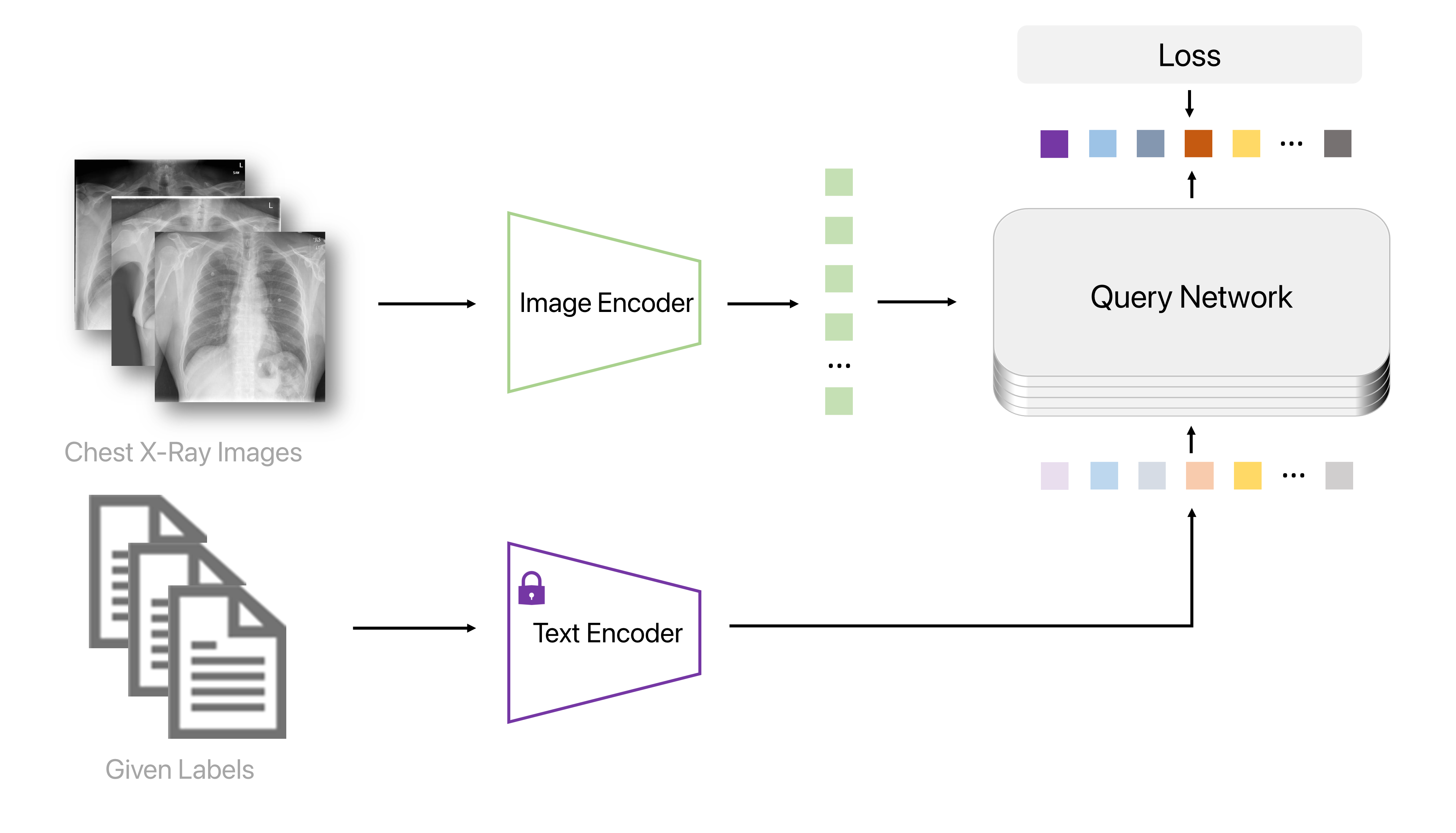}
    \caption{The general model framework of our solution. Two modality-specific encoders first generate image and label features. Then the ground truth labels help supervise the prediction of textually enhanced image feature after multiple transformer decoder layers. 
    }
    \label{fig:architecture}
\end{figure*}

The problem of interest on chest X-rays (CXRs) can be formalized as a multi-label classification problem. Let $\mathcal{X}$ be the input space and $\mathcal{Y}$ be the class space. The training set can be denoted as $\mathcal{D} = \{(x_i, Y_i)\}_{i=1}^N$, where any input CXR image $x_i \in \mathcal{X}$ is associated with a label set $Y_i\subset \mathcal{Y}$ of clinic findings in $x_i$. Our goal is to train a deep model $f$ on the training set $\mathcal{D}$, which predicts the likelihood that any class $y \in \mathcal{Y}$ exists in any CXR image $x\in \mathcal{X}$, \textit{i.e.}, $f: \mathcal{X}, \mathcal{Y} \rightarrow [0,1]$. Different from the traditional supervised learning paradigm that translates the image features to fit the 0-1 label vector, we also incorporate the pretrained text encoders to extract the class embedding that has incorporated the label correlation. Specifically, the model prediction for any image $x$ over all categories can be denoted as
\begin{equation}
    S(x) = \Phi_{\text{query}}(\Phi_{\text{image}}(x), \Phi_{\text{text}}(\mathcal{Y})),
\end{equation}
where $S(x)\in [0,1]^{\lvert\mathcal{Y}\rvert}$ refers to the prediction of the image $x$ and all diseases in $\mathcal{Y}$. $\Phi_{\text{image}}(\cdot)$, $\Phi_{\text{text}}(\cdot)$, and $\Phi_{\text{query}}(\cdot)$ are the image encoder, the text encoder, and the query network, respectively. Here, we employ cross-entropy loss as the guiding criterion for training the model. The whole architecture is illustrated in Fig.~\ref{fig:architecture}.


\textbf{Image Encoder.} For a given CXR image $x \in \mathbb{R}^{H \times W \times 3}$, its visual features $\Phi_{\text{image}}(x) \in \mathbb{R}^{h \times w \times d}$ can be obtained by a visual feature extractor, where $h, w$ and $d$ refer to the height, width, and feature dimension of
the output feature map. The image encoder can be almost any well-known vision models, \textit{e.g.}, ResNet-50~\cite{he2016deep} and DenseNet-121~\cite{huang2017densely}. Here we preserve the visual features before pooling that can provide richer information in the subsequent query network.


\textbf{Text Encoder.} Following the spirit of prevalent pretraining techniques, we construct the text encoder to compute the semantic features $\Phi_{\text{text}}(\mathcal{Y})\in \mathbb{R}^{\lvert\mathcal{Y}\rvert \times d}$ for all the category texts in $\mathcal{Y}$, where $d$ is the dimension of textual feature. To sufficiently characterize the semantic, the text encoder is pre-trained on the medical textual data. During training, we freeze the text encoder to avoid overfitting.

\textbf{Query Network.} After extracting the features of the two modalities, we use multiple transformer decoder layers for disease diagnosis, with label embedding $\Phi_{\text{text}}(\mathcal{Y})$ as Query, and CXR image feature $\Phi_{\text{image}}(x)$ as Key and Value. The transformer decoder aims to combine textual information into corresponding image features, further improving predictive performance. The outputs of the sequential transformer decoder layers are further fed into one MLP layer.

\begin{table*}[t]
  \centering
  \caption{Results of some different combinations of designs on the development set.}
  \begin{tabular}{cccccc|c}
    \toprule
    Image Encoder & Text Encoder & Separate Classifier & Reweighting & MixUp & TTA & mAP\\
    \midrule
     ResNet-50 & PubMedBERT& & & & & 0.3187\\
     ResNet-50& PubMedBERT&\checkmark & & &  & 0.3193\\
     ResNet-50& PubMedBERT&\checkmark & & &   \checkmark & 0.3220\\
     ResNet-50 & PubMedBERT&\checkmark & \checkmark, 2 & &  \checkmark & 0.3273\\
     DenseNet-121 & PubMedBERT&\checkmark & \checkmark, 2 & &  \checkmark & 0.3236\\
     ResNet-50& Clinical-T5 &\checkmark & \checkmark, 3 & \checkmark  & \checkmark & 0.3273\\
     ResNet-50& PubMedBERT&\checkmark & \checkmark, 3 & \checkmark &   \checkmark & 0.3247\\
     ResNet-50& Clinical-T5 &\checkmark & \checkmark, 2 & \checkmark  & \checkmark & 0.3244\\
     ResNet-50& PubMedBERT&\checkmark & \checkmark, 2 & \checkmark & \checkmark & \textbf{0.3280}\\

    \bottomrule
  \end{tabular}
  \label{tab1}
\end{table*}

\subsection{Advanced Designs}\label{advanceddesigns}

On the basis of the above general framework, we introduce a series of detailed designs from seven different perspectives in the following, from first steps like feature extraction to the regular techniques during the inference time.

\textbf{Textual Feature Extractor.} For the extraction of label embedding, we utilize two off-the-shelf pre-trained medical domain-specific encoders respectively. The first is PubMedBERT~\cite{zhang2023knowledge} which was finetuned on one Unified Medical Language System~(UMLS) knowledge graph~\cite{bodenreider2004unified}, the second is Clinical-T5~\cite{lehman2023clinical} which was a Large Language Model built with MIMIC clinical texts. 

\textbf{Loss Function Reweighting.} 
In the conventional multi-label learning framework, the loss function is averaged by sample across the entire dataset, resulting in a model that is heavily biased towards the head classes and performs poorly in the tail classes.
Following the spirit of reweighting~\cite{menon2013statistical, morik1999combining} in imbalanced learning and learning from failure~\cite{suh2019stochastic, tang2019uldor} in hard example mining, we reweight the loss of those classes that performed poorly on the development set. Then, we can adjust the training bias towards the head classes, and force the importance of the tail classes. 

\textbf{Separate Classifier.} For the last linear layer of the query network, we assign class-wise classifiers for each given class to preserve the independence of predictions between classes, with differs from the common practice that all classes share one classifier. In this way, the impact of the up-weighting of the tail and hard categories on the other categories can also be reduced.

\textbf{Data Augmentation.}  MixUp~\cite{zhang2018mixup} is a common data augmentation technique that directly mixes up different images and labels in a batch. According to a Beta distribution, the hyper-parameter $\alpha$ regulates the shape of the distribution, thus modulating the intensity of MixUp. We apply the MixUp to augment the sample space to improve the generalization performance of the model.

\textbf{Test-time Augmentation (TTA).} 
We conduct test-time augmentation for test images. Multiple transformations are applied on the test image, such as random cropping, scaling, flipping, etc.. Then, we test the model on transformed images, and take the average prediction as the output, which improves the stability and accuracy of the prediction. 

\textbf{Class-Wise Ensemble.} The multi-label nature allows us to train all categories simultaneously in order to leverage the dependency between classes, but to predict each category independently when testing. For each class, we select best models based on their performance on the development set and average their predictions, further improving the performance. Note that, it is different from the ordinary way that selects the candidates for ensemble based on the model performance instead of each class performance.

\textbf{Exogenous Data Replenishment.} Apart from the given training set, we also train the model together with other CXR datasets like ChestXRay14~\cite{wang2017chestx} and CheXpert~\cite{irvin2019chexpert}. We choose the official training set of both datasets for model training and the label space of both datasets is 14. For those labels the competition specify while not mentioned in these two datasets, we set them to be $0$, {\em i.e.}, assuming corresponding diseases don't exist. Actually, this can be further improved from the perspective of semi-supervised learning.
\section{Experiments}

\subsection{Datasets}
The ICCV
CVAMD 2023 CXR-LT Competition is built on an expanded version of \textbf{MIMIC-CXR-JPG}~\cite{johnson2019mimic, holste2022long}, which is a benchmark dataset for chest X-ray classification and contains 377k CXR images collected from over 227k studies. Except 14 clinical findings already annotated, the expanded version includes 12 new annotations of rare disease extracted from radiology reports following~\cite{holste2022long}.

\textbf{ChestXRay14}~\cite{wang2017chestx} has 112,120 CXR images released by NIH while \textbf{CheXpert}~\cite{irvin2019chexpert} has 224,316 CXR images collected from 65,240 patients. We utilize them as the source of external data supplement in the manner stated in Sec.~\ref{advanceddesigns}.

\subsection{Settings}
At the training stage, images are resized to $512\times 512 \times 3$. Some regular data augmentation methods, such as resize crop, horizontal flip, rotation, are randomly applied to the process of data preprocess. For the extraction of image feature, we take the output from the 4th residual block of the ResNet-50~\cite{he2016deep} or the final output of DenseNet-121~\cite{huang2017densely}, the output after visual encoder is further fed into MLP layers, to obtain the same feature dimension $d = 768$ as that of the textual feature. After each modality-specific encoder, we adopt $4$ transformer decoder layers for the subsequent operations. The $\alpha$ of the Beta distribution in MixUp is set to be $4$. We upweight the loss of nine classes performed poorly on the development set, {\em i.e.}, \textit{calcification of the aorta}, \textit{fibrosis}, \textit{infiltration}, \textit{lung lesion}, \textit{pleural other}, \textit{pleural thickening}, \textit{pneumomediastinum}, \textit{pneumoperitoneum}, \textit{tortuous aorta}, and the upweighting factor is $2$ or $3$. The merging type of TTA is geometric-mean.

For all experiments, we use AdamW~\cite{loshchilov2017decoupled} optimizer with cosine annealing scheduler, together with learning rates of $5\times 10^{-5}$ for regular training and $1\times 10^{-6}$ for warm up.
We train the model on NVIDIA A100s or RTX 3090s with batch size $32$ for up to $50$ epochs. The first $20$ epochs are set for warming up.

\begin{table}[t]
  \centering
  \caption{Results of model ensemble and data replenishment on the test set.}
  \begin{tabular}{cc|c}
    \toprule
     Ensemble Method & Data Replenishment & mAP\\
    \midrule
     
    \textit{Model-wise Ensemble} & & 0.347\\
    \textit{Class-wise Ensemble} & & 0.348\\
    \textit{Class-wise Ensemble} & \checkmark& \textbf{0.349}\\

    \bottomrule
  \end{tabular}
  \label{tab2}
\end{table}

\subsection{Results}

During the development phase, we conduct experiments according to the combinations of designs in Sec.~\ref{advanceddesigns}, containing the selection of the two modality-specific encoders, whether adopt separate classifier, reweighting, MixUp, TTA or not respectively, different parameters of reweighting~($2$ or $3$). From Tab.~\ref{tab1}, we can see that when separate classifier, reweighting, MixUp and TTA are simultaneously combined together with ResNet-50 and PubMedBERT, the best mAP score is achieved on the development set. Apparently, the incorporation of a series of LT-specific designs help boost the chest X-rays disease diagnosis ability of the base general architecture from $0.3187$ to $0.3280$.

In addition to exploring the design of individual models, we also investigate the performance of model ensembles and exogenous data replenishment. During the final submission stage (test phase), we rank the performance of individual models based on their development set scores and leverage the model ensemble. Specifically, we use two ensemble methods: \textit{model-wise ensemble}, where we directly average the predictions of distinct models, and \textit{class-wise ensemble}, as described in Sec.~\ref{advanceddesigns}. As demonstrated in Tab.~\ref{tab2}, the \textit{class-wise ensemble} outperforms the \textit{model-wise ensemble}, and the \textit{class-wise ensemble} altogether with data replenishment achieves the highest score among all attempts at a remarkable $0.349$ mAP.

\section{Conclusion}
This paper presents our solution to the ICCV CVAMD 2023 CXR-LT Competition. To tackle the challenging long-tailed multi-label classification problem in chest X-rays (CXRs), we explore various techniques from different perspectives. Through experiments on the provided dataset, we demonstrate the effectiveness of our proposed framework. In future work, we plan to investigate visual-language pre-training~(VLP) in medical applications to further enhance the capacity of AI-aided CXR disease diagnosis.


{\small
\bibliographystyle{ieee_fullname}
\bibliography{egbib}
}

\end{document}